\documentclass[graybox]{svmult}
\usepackage{gensymb}

\usepackage{type1cm}        
\usepackage{makeidx}         
\usepackage{graphicx}        
\usepackage{multicol}        
\usepackage[bottom]{footmisc}

\usepackage{newtxtext}       %
\usepackage{newtxmath}       

\usepackage[section]{placeins}

\usepackage[normalem]{ulem}
\usepackage{subcaption}
\makeindex             

\begin{document}

\title*{Challenges in Plane Symmetry: From Theory to Perception.}
\author{Furkan Çengel, Venera Adanova and Sibel Tari}
\institute{Furkan Çengel \at Middle East Technical University, 06800 Ankara, Turkey, \email{furkancengel@gmail.com }
\and Venera Adanova \at TED University, 06420 Ankara,Turkey, \email{venera.adanova@tedu.edu.tr}
\and Sibel Tari \at Middle East Technical University, 06800 Ankara, Turkey, \email{stari@metu.edu.tr}}
%
%
\maketitle

\abstract*{ }

\abstract{The planar ornaments are created by repeating a base unit using a combination of four primitive geometric operations: translation, rotation, reflection, and glide reflection. According to group theory, different combinations of these four geometric operations lead to different symmetry groups. In this work, we select a single challenging ornament, and analyse it both from the theoretical point of view and perceptual point of view. We present the perceptual experiment results, where one can see that the symmetries that the participants perceived from the ornaments do not match to what the theory dictates.    }

\section{Introduction}
\label{sec:1}
 One of the problems in shape analysis is to find the symmetries in a given shape. A single shape might contain mirror reflectional and rotational symmetries. These symmetries are defined as  distance-preserving transformations (isometric operations) of some object that leave that object unchanged.
 Detection of these isometries in a given shape is a challenge in itself. But, what if we copy this very shape and repeat it infinitely many times in two directions? Then we would have four isometric operations: translation, rotation, mirror reflection, and glide reflection. By repeating a shape using different combinations of these isometric operations we can produce different kinds of ornaments. Depending on the isometries used in the creation of the ornament we define the symmetry group of it. There are exactly seventeen symmetry groups, meaning that there are seventeen different ways to fill the plane by repeating a given shape via the combinations of these four isometries. This fact has been known for nearly a century after the works published by~\cite{Polya,Niggli24,Niggli26}. These seventeen symmetry groups are known as $Wallpaper$ $Groups$ and are listed as $p1$, $pm$, $pg$, $cm$, $p2$, $pmm$, $pmg$, $pgg$, $cmm$, $p3$, $p3m1$, $p31m$, $p4$, $p4m$, $p4g$, $p6$, and $p6m$. Each character in the group names has a meaning. Thus, the digit that follows letter $p$ or $c$ indicates the highest order of rotation, and the letters $m$ and $g$ stand for mirror and glide reflection, respectively. The letter $p$ is used for $primitive$ $cell$ and the letter $c$ is for $centered$ $cell$ structures.
 
\begin{figure}[!htb]
  	\centering
  	\begin{tabular}{cp{0.2cm}cp{0.2cm}cp{0.2cm}c}
  		\includegraphics[width=0.2\linewidth]{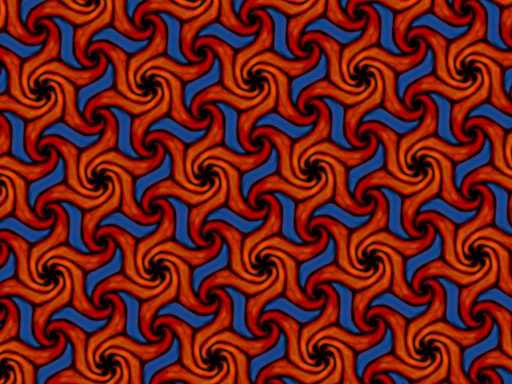}&&
  		\includegraphics[width=0.2\linewidth]{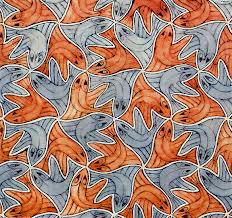}&&
  		\includegraphics[width=0.19\linewidth]{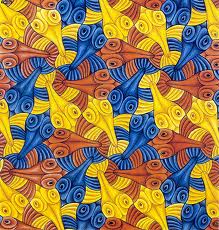}&&
  		\includegraphics[width=0.2\linewidth]{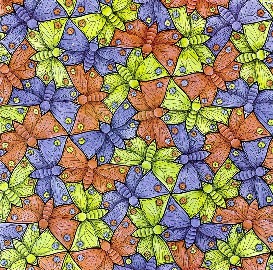}\\
  		
  		\includegraphics[width=0.2\linewidth]{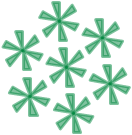}&&
  		\includegraphics[width=0.2\linewidth]{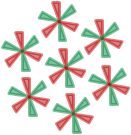}&&
  		\includegraphics[width=0.2\linewidth]{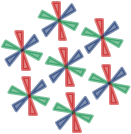}&&
  		\includegraphics[width=0.2\linewidth]{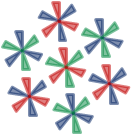}\\
  		(a)&&(b)&&(c)&&(d)\\
  	\end{tabular}
  	\caption{ Ornaments that belong to the same symmetry group in an uncolored case. When the color symmetry is allowed all four belong to different groups. The second row illustrates the generation rules for the ornaments given in the first row. }	\label{fig:p6}
  \end{figure}

  The seventeen symmetry groups do not consider color permutations that lead to color symmetry. An ornament has color symmetry, if applying certain isometric operation maps all regions of one color to the regions of another color. The shapes of mapped regions should be identical. When the color symmetry is involved, the number of groups that can be produced by filling the plane depends on the number of colors being used. For example, for these seventeen symmetry groups the number of all possible colorings using two colors equals to $46$, and equals $23$ when three colors are used. An example for the color symmetry groups is illustrated in Fig.~\ref{fig:p6}. First ornament (Fig.~\ref{fig:p6}(a)) has no color symmetry. If the color permutations are ignored, all four ornaments belong to group $p6$. When we consider colors, the second ornament (Fig.~\ref{fig:p6}(b)) is assigned to one of the groups of possible colorings using two colors of an ornament with underlying symmetry group $p6$, i.e. $p6/p3$. Fig.~\ref{fig:p6}(c) belongs to one of the groups of possible colorings using three colors of an ornament with underlying symmetry group of $p6$ ($p6/p2$). The $mariposa$ given in Fig.~\ref{fig:p6}(d) represents a color-symmetry-imperfect coloring of a form-wise perfect construction.The second row  in Fig.~\ref{fig:p6} shows the construction rules applied for the ornaments. Note that the underlying symmetry group of all four ornaments is $p6$.
  
Typically, the study of ornaments is based on classical mathematical modeling via group theory. Here, the aim is to find the basic repeating element in an ornament, detect the symmetries applied in order to repeat this element, and associate an ornament with some predefined group. A computer implementation for the group theoretic approach can be found in~\cite{Adanova2019, Liu2004_pami}, where ornaments are analysed and  classified into symmetry groups depending on detected isometries. However, computational models follow straightforward rules and fail to understand human perception.

One of the earliest works emphasizing the importance of symmetry in perception can be found in the works of Attneave~\cite{Attneave54, Attneave55}. He argued that symmetric figures are easier to reproduce than asymmetric ones because the symmetric figures, as they consist of repeating units, contain less information, making them more memorable and visually pleasing due to their simplicity.

 Several studies on the perception of symmetries aimed to understand people's ability to differentiate between properties of symmetry. Washburn et  al. ~\cite{washburn1988symmetries} suggest that people recognize two-fold rotational symmetries better than four-fold or eight-fold symmetries. The results of an experiment in which participants grasped two-fold rotational symmetries that contain a vertical or an oblique axis more comfortably than ones that have a horizontal axis are reported in ~\cite{corballis1974perception}. Vertical symmetry was the least complicated symmetry type to be detected by participants in~\cite{palmer1978orientation}. Shepard et al.~\cite{shepard1971mental} explain these results; to perceive the symmetry, people rotate symmetrical structures in their minds until the axis of symmetry is vertical.  

Landwehr~\cite{landwehr2011visual} experiments with four wallpaper groups (p1, p2, pm, and pg), and observes that two-fold rotations are better distinguished than mirror and glide reflections. The study also argues that wallpaper groups are not in one-to-one relationships with their unit cells and fundamental domains –generating regions of the repetitive structure– and this ambiguity affects the perception of symmetry, hurts the chance of discrimination between wallpaper groups solely on the implied structure of repetitiveness.

In~\cite{Clarke} Clarke et al. conducted an experiment where the participants had to sort the ornaments according to their symmetry groups.The results show that the ornaments that are in the same wallpaper group are classified together by humans more often than would be expected by chance. Along with this result, the authors also conclude that human classification depends on something else, besides wallpaper groups, which is not being captured by computational models.

In their recent work Kohler et al. ~\cite{kohler2021} conducted an experiment by collecting both brain imaging and behavioral data from participants using all wallpaper groups. Their results show that more complex groups, groups that contain more isometries, produce larger response in human visual cortex and lead to shorter symmetry detection thresholds.

\begin{figure}[!htb]
	\centering
	\begin{tabular}{cp{1cm}c}
		\includegraphics[width=0.3\linewidth]{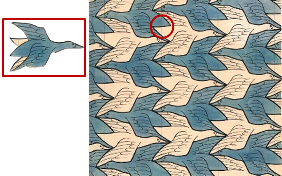}&&
		\includegraphics[width=0.2\linewidth]{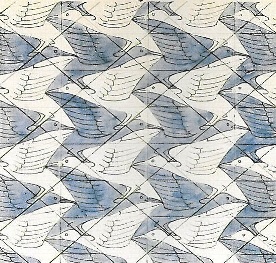}\\
		(a)&&(b)\\
	\end{tabular}
	\caption{Two quite similar ornaments produced using two different repetition rules. (a) An ornament of $p1$ group. Produced by translating two distinct figures, which are similar. The red box contains an image where the white and blue birds are superimposed. It shows the distinction between these two shapes. (b) An ornament of $pg$ group. Contains one bird figure of two different colors. A bird of one color is a glide reflection of a bird of another color. }
	\label{fig:birds}
\end{figure}

There are some challenging cases where the computational models fail. One of these cases is illustrated in Fig.~\ref{fig:birds}. The first ornament contains two different birds that are translated in two directions to fill the plane. The symmetry group of this ornament is $p1$. The second one contains one bird in two different colors that are related by glide reflection. Hence, its symmetry group is $pg$. However, those two images are perceptually very close to each other. The red box in  Fig.~\ref{fig:birds}(a) contains an image where the blue and white birds are superimposed.  While the tail of a blue bird looks downward, the tail of the white bird looks upward. Otherwise, those two birds would be related by glide reflection, which would imply that two ornaments in Fig.~\ref{fig:birds} were created using the same production rules. By adding a minor change to the first image, the artist broke the symmetry. 

Indeed, there are many example ornaments that seem to have higher symmetry, but thorough analyses reveal that the artist has introduced some tricky part that breaks this higher symmetry.
In this work, we focus on a single challenging ornament. We perform theoretical analysis of this ornament in the next section. Then we present some survey results, which are also built around this ornament. 

\section{Moroccan Ornament}
\label{sec:2}

\begin{figure}[!htb]
\centering
\begin{tabular}{c}
\includegraphics[width=0.2\linewidth]{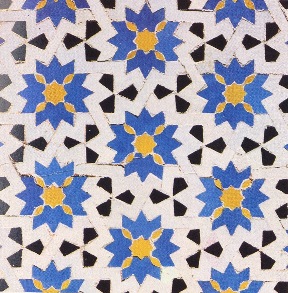}
\end{tabular}
\caption{Moroccan Ornament}\label{fig:original}
\end{figure}

Consider the Moroccan ornament given in Fig.~\ref{fig:original}. At first glance, it seems that the ornament is rich for symmetries. There are six-fold rotations surrounded by three-fold rotations. The six-fold rotations are obtained by blue dodecagrams, while the black three-leaved shapes form three-fold rotations. A perfect example for an ornament of $p6m$ group.  Actually, this ornament is quite more challenging than it looks.

\begin{figure}[!htb]
\centering
\begin{tabular}{ccc}
\includegraphics[width=0.55\linewidth]{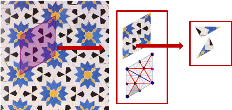}&
\includegraphics[width=0.15\linewidth]{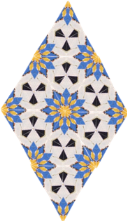}&
\includegraphics[width=0.15\linewidth]{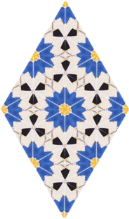}\\
(a)&(b)&(c)\\
\end{tabular}
\caption{Generation of $p6m$ ornament using two different fundamental domains (a) Obtained UC assuming that an ornament has 6-fold rotations and extracted two FDs. (b)-(c) Generated $p6m$ ornaments using two different FDs.}\label{fig:p6m_gen}
\end{figure}

Assuming that the ornament in Fig.~\ref{fig:original} belongs to $p6m$ group we extract a unit cell (UC), which is the smallest unit of the ornament translating which we can reconstruct the original ornament. Note that the unit cells that we obtain in this analysis part are all proper unit cells, meaning that they obey all the rules of the respective symmetry groups. Hence, using a unit cell that we extract a whole pattern can be generated either by translating or applying the symmetry group rules of the corresponding ornament. Thus, if the Moroccan ornament indeed has six-fold rotations on blue dodecagrams than according to the rules of $p6m$ group the proper unit cell should be exactly the one shown in Fig.~\ref{fig:p6m_gen} (a), so that the corners of the unit cell reside on six-fold rotation centers. The proper unit cell for $p6m$ group is shown in Fig.~\ref{fig:p6m_gen} (a) (bottom image in a red box). Note that, since we have only one type of six-fold rotations then the proper unit cell for $p6$ and $p6m$ groups are unique.  From the unit cell one can obtain the fundamental domain (FD), the smallest unit of the ornament applying on which the symmetry group rules enable the reconstruction of the original ornament. As was mentioned earlier, for $p6m$ group the proper unit cell is unique, then the fundamental domain must also be unique. The fundamental domain is shown as a gray region in Fig.~\ref{fig:p6m_gen} (a) (bottom image in a red box). The red and blue lines in the  proper unit cell represent mirror reflections. Hence, all small triangular regions are mere mirror reflections of each other. It means that, instead of the gray triangle, one can choose any of these small triangles as the FD, since they are all similar. However, we obtain two different FDs from the Moroccan ornament. Fig.~\ref{fig:p6m_gen}(a) illustrates the extracted UC and two different FDs. Using each of these FDs and applying the rules of $p6m$ group we generate two different ornaments as shown in Fig.~\ref{fig:p6m_gen}(b) and (c). None of them match the original ornament. Hence, the ornament is not of $p6m$ group.

\begin{figure}[!htb]
\centering
\begin{tabular}{cccc}
\includegraphics[width=0.2\linewidth]{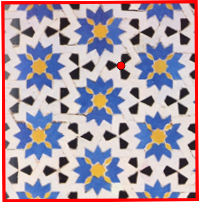}&
\includegraphics[width=0.3\linewidth]{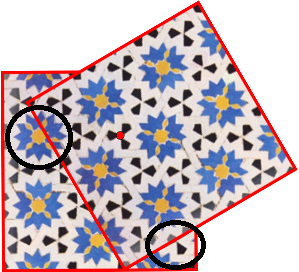}&
\includegraphics[width=0.2\linewidth]{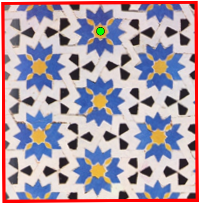}&
\includegraphics[width=0.28\linewidth]{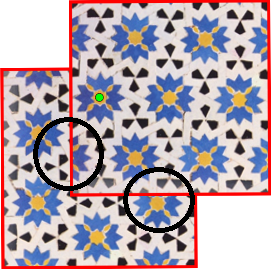}\\
(a)&(b)&(c)&(d)\\
\end{tabular}
\caption{Illustrating three-fold and four-fold rotations. (a) The center of three-fold rotation shown as red dot. (b) Rotation for $120 \degree$. Rotated image is laid on top of the original image. Observe the regions marked with black circles. These are the obvious mismatch places. (c) The center of four-fold rotation shown as green dot. (b) Rotation for $90 \degree$. Rotated image is laid on top of the original image. Obvious mismatch places marked as black circles. }\label{fig:moroccoRots}
\end{figure}

 More detailed analyses show that the yellow leaves extending from the yellow octagram have broken the symmetry. Now its clear that there are four-fold rotations instead of six, surrounded by three-fold rotations. According to the symmetry group theory four-fold and three-fold rotations never come together in an ornament. We check if an ornament has any of these symmetries. The supposedly three-fold rotation center should lie at the centers of black three-leaved shapes. We select a point as shown in Fig.~\ref{fig:moroccoRots}(a) and rotate a copy of an ornament around that point for $120\degree$. If there is actually a three-fold rotation around that point then the rotated image should perfectly fit the original image. Hence, the rotated image would just naturally continue the original image's motifs without abrupt changes and distortions. Observe that the yellow leaves of the rotated image, which is laid on top of the original image, do not fit the yellow leaves of the original image (Fig.~\ref{fig:moroccoRots}(b)), indicating the absence of three-fold rotation. Obvious mismatches are marked by black circles in Fig.~\ref{fig:moroccoRots} (b).  We then check for four-fold rotation by selecting the center of yellow octagram as four-fold rotation center (Fig.~\ref{fig:moroccoRots}(c)). Rotating an ornament around the green point shows (Fig.~\ref{fig:moroccoRots}(d)) that the ornament has no four-fold rotation either. Observe that while the rotated image should flawlessly continue the motifs of original one, the abrupt motif changes are seen in the places where the original and rotated images are connected.  

\begin{figure}[!htb]
\centering
\begin{tabular}{ccc}
\includegraphics[width=0.15\linewidth]{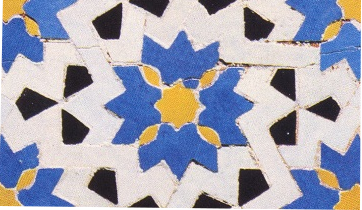}&
\includegraphics[width=0.2\linewidth]{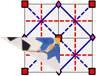}&
\includegraphics[width=0.2\linewidth]{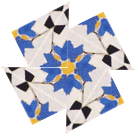}\\
(a)&(b)&(c)\\
\end{tabular}
\caption{(a) UC extracted from the ornament assuming that the ornament belongs to $p4g$ group. (b) Corresponding FD. (c) Generated ornament using FD and the rules of $p4g$ group.}\label{fig:p4g_gen}
\end{figure}

Fig.~\ref{fig:p4g_gen}(a)-(b) shows possible UC and FD assuming that the ornament has four-fold rotation. Since one can observe only one type of four-fold rotation centers, the ornament would belong to $p4g$ group. According to the rules of $p4g$ symmetry group the proper unit cell should obey the symmetries given in Fig.~\ref{fig:p4g_gen}(b). Hence, the unit cell corners should reside on four-fold rotation centers and there should be one four-fold rotation center in the middle of the unit cell. Observe that the generated ornament (Fig.~\ref{fig:p4g_gen}(c)) is different from the original one.

\begin{figure}[!htb]
\centering
\begin{tabular}{cccc}
\includegraphics[width=0.3\linewidth]{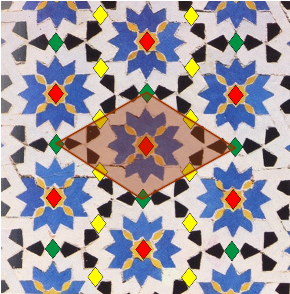}&
\includegraphics[width=0.17\linewidth]{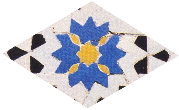}&
\includegraphics[width=0.1\linewidth]{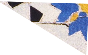}&
\includegraphics[width=0.3\linewidth]{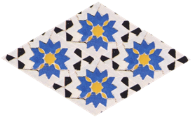}\\
(a)&(b)&(c)&(d)\\
\end{tabular}
\caption{(a) Symmetries detected for ornament. Rhombi indicate two-fold rotation centers. (b) Extracted UC. (c) Corresponding FD. (d) Generated ornament.}\label{fig:moroccoSym}
\end{figure}

We continue our meticulous analyses and find out that the maximum order of rotation that the ornament has is two. Two-fold rotations of the ornament are illustrated in Fig.~\ref{fig:moroccoSym}(a) via rhombi of different colors.  While the green and yellow rotation centers seem to be similar, there is a subtle distinction between them. There are mirror reflection lines passing through the green centers, while the yellow rotation centers are the intersection points of glide reflection axes. We detected three different two-fold rotation centers. Among seventeen symmetry groups only the $cmm$ group has three different two-fold rotation centers. And the proper unit cell is the one which has one type of two-fold rotation centers (green rhombi) on the corners, second type of two-fold rotation centers (yellow rhombi) on the unit cell edges, and the third type of two-fold rotation (red rhombi) is at the center of the unit cell. Observe that no other unit cell would obey these rules than the one that is depicted in Fig.~\ref{fig:moroccoSym}. Thus, one cannot put the red two-fold rotation centers on the unit cell corners, because in that way the unit cell edges would have two different types of two-fold rotation centers. Hence, the unit cell that we obtained is unique, so is the corresponding FD. Recall that these rules are important only to obtain proper unit cell. But if one is interested only in translational unit cell then it is not necessarily unique.  Fig.~\ref{fig:moroccoSym}(b) and (c) depict the extracted UC and FD using the detected symmetries in Fig.~\ref{fig:moroccoSym}(a). Finally, Fig.~\ref{fig:moroccoSym}(d) shows the generated ornament using the extracted FD and applying the generation rules of $cmm$ symmetry group. Observe that the generated ornament matches the original ornament. Thus, the ornament belongs to $cmm$ group.

\section{Experiments}
\label{sec:3}
Analysis of the symmetries of the Moroccan ornament revealed that it belongs to $cmm$ group. It takes  detailed analyses for one to detect the symmetry group of this ornament, since at first sight the Moroccan ornament seems to have higher order symmetries (like six-fold rotations). In the following section we present two perceptual experiment results, where the Moroccan ornament plays the key role. Our aim is to understand what symmetries do the people see in this ornament.

\subsection{Dataset}

\begin{figure}[!htb]
	\centering
	
	\includegraphics[width=0.99\linewidth]{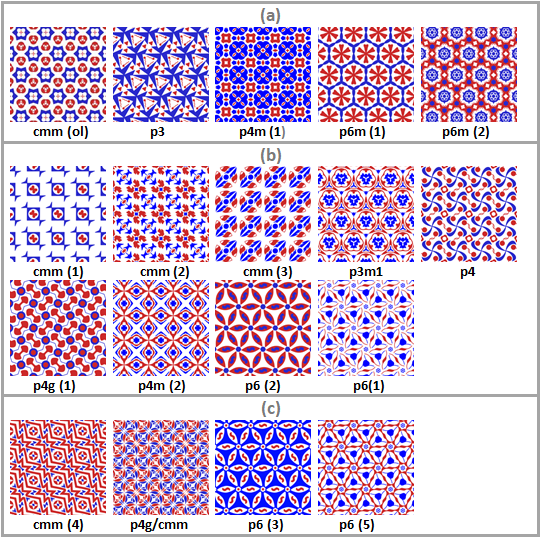}\\
	
	\caption{Dataset. There are three parts. Part $a$ was used in both experiments. Part $b$ was used in the first experiment only, while part $c$ was used in the second experiment only.}\label{fig:dataset}
\end{figure}

We have created $18$ ornaments, of different symmetry groups using iOrnament tool~\cite{iornament}, for the experiments. The ornaments are all drawn with three colors: red, white and blue. The dataset is divided into three parts as shown in Fig.~\ref{fig:dataset}. Part $a$ contains ornaments that were used in both experiments, part $b$ was used in the first experiment only, while part $c$ was used in the second experiment only. The dataset contains one sample from each of the $p3$, $p3m1$, $p4$ and $p4g$ groups, two samples from $p4m$ group, four samples from $p6$ group and two samples from $p6m$ group. There are five ornaments from $cmm$ group, and the only ornament with color symmetry of $p4g/cmm$ group. 

A special attention needs the first ornament in Fig.~\ref{fig:dataset} (part (a)), which is labelled as $cmm$ $(ol)$. This ornament was created by overlapping two symmetry groups: first, two different ornaments of $p6m$ and $p4m$ groups were created and then combined together in order to get an ornament similar to Moroccan ornament. In this manner, we deliberately broke the symmetry of the ornament of $p6m$ group, just like it was done by the Moroccan artist. Note that most of the ornaments in the dataset belong to $cmm$ group, and all the other ornaments contain the symmetries that the Moroccan ornament seems to have, like four-fold rotations, three-fold rotations, etc. 

\subsection {First Experiment}

\begin{figure}[!htb]
	\centering

	\includegraphics[width=0.95\linewidth]{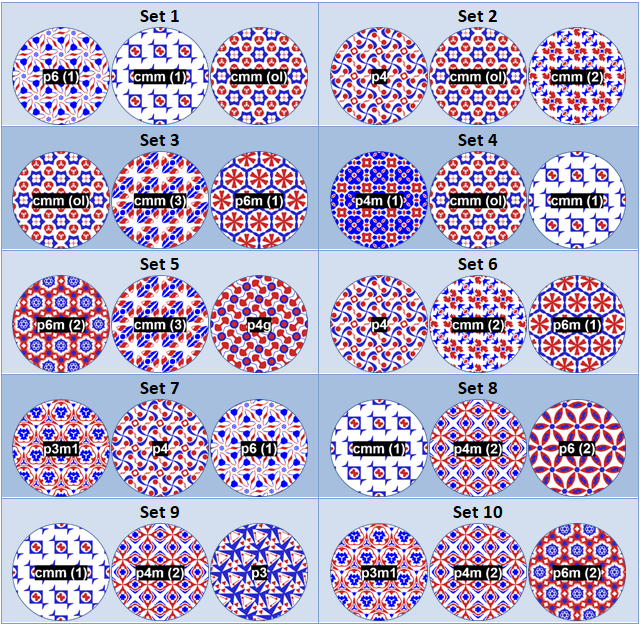}\\
	
	\caption{Ornament sets for the first experiment}\label{fig:set1}
\end{figure}

For the first experiment, we used the ornaments from parts $a$ and {b} of the dataset. From these $14$ ornaments we constructed $10$ small ornament sets so that each set contains three ornaments. The ornament sets are illustrated in Fig.~\ref{fig:set1}. For each set in Fig.~\ref{fig:set1} two tasks were designed: finding the most similar and least similar ornament to the query ornament.  The query ornament is the Moroccan ornament in all tasks. Thus, overall $20$ tasks were designed. For the first $10$ tasks participants are required to select the most similar ornament to the Moroccan ornament among three options, while for the last $10$ tasks participants are asked to select the least similar ornament.

\begin{figure}[!htb]
	\centering
	
	\includegraphics[width=0.8\linewidth]{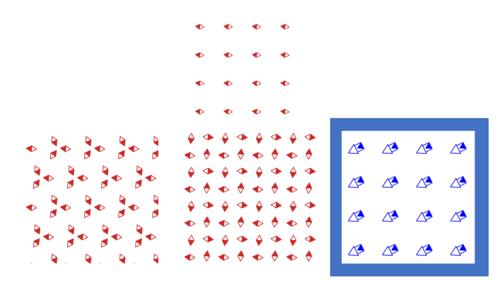}\\
	
	\caption{Warm up task for the first experiment. A query ornament is given on top. The participants are asked to find the ornament with similar repetition structure among three options. Was designed deliberately to show that colors and shapes are not important when considering ornament similarity.}
	\label{fig:warmup1}
\end{figure}

$30$ participants took part in this experiment. All participants are students of Computer Engineering Department. Before beginning the experiment participants had a brief introductory slides describing isometric operations used to create ornaments. An animation was shown where an ornament of $p6$ group was generated starting from a fundamental domain. Note that the symmetry groups were not discussed at all. We emphasised the importance of repetition structure. Thus, one warm up task (Fig.~\ref{fig:warmup1}) was shown to participants to illustrate that colors and shapes are not to be considered. In this task we asked participants to select among three ornaments the one with the similar repetition structure as query ornament. Two of the three ornaments given as options were created using the same motif and color as query ornament, but they had different repetition structures. The third ornament was created using different motif and color, but had similar repetition structure as the query ornament. Participants were given some time to decide on the answer, then the correct answer was revealed and repetition structures of each ornament were discussed.  Afterwards we proceeded with real experimental tasks. The tasks were shown on slides, while participants wrote their answers in a form. The participants were given thirty seconds for each task.

\begin{figure}[!htb]
	\centering
	
	\includegraphics[width=0.90\linewidth]{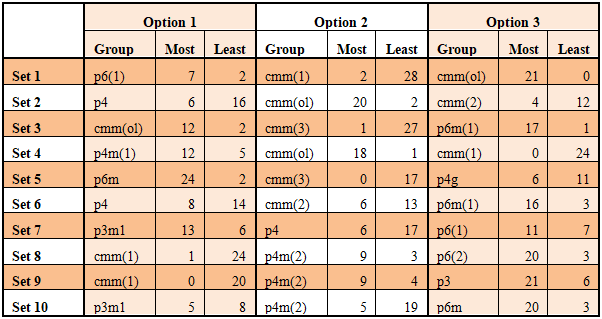}\\
	
	\caption{The result for the first experiment.}\label{fig:set1_result}
\end{figure}

The results of the first experiment can be seen in Fig.~\ref{fig:set1_result}. For each ornament set we combined the results for both tasks, i.e. the number of times that a particular ornament from the options was selected to be most similar and least similar. For example, for the first set the ornament of $p6$ group was selected by $7$ participants as the most similar one, while $2$ participants thought that it was least similar to Moroccan ornament.

From the participants’ answers, we construct distance matrices that show how consistent are the participants between each other for each set and task. The distance between the answers of two participants is computed using the Kendall tau distance ~\cite{kendall}, which is used to measure the disagreement between two ranking lists. First, the answers are scored as 1 for the most similar ornament, 3 for the least similar ornament, and 2 for the unselected ornament. Then, the Kendall tau distance between two ranking lists is defined as

\begin{equation} \label{eq:kendall}
K(\tau_1, \tau_2)=\sum_{i,j\in{P}}{\bar{K}_{i,j}\left({\tau_{1}, \tau_{2}}\right)}
\end{equation}

where $P$ is the set of pairs of distinct elements in ranking lists $\tau_1$ and $\tau_2$, and $\bar{K}_{i,j}\left({\tau_{1}, \tau_{2}}\right)$ is the penalty function for $i$ and $j$, which is $0$ when they are in the same order and 1 when they are in the opposite order. For the distance matrices, distance values are normalized as in (\ref{eq:normalized_kendall}), where $n$ is the number of ornaments in each set.

\begin{equation} \label{eq:normalized_kendall}
\frac{K(\tau_1, \tau_2)}{\frac{n(n-1)}{2}}
\end{equation}

\begin{figure}[!htb]
	\centering
	
	\includegraphics[width=0.99\linewidth]{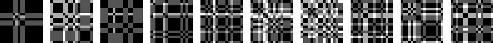}\\
	
	\caption{17x17 Distance Matrices, obtained for ten sets, after excluding inconsistent participants.}\label{fig:dt}
\end{figure}

Some participants selected the same ornament for both least similar and most similar cases. These participants are labelled inconsistent and are not considered in distance matrix computation; 13 participants had at least one inconsistent answer. Distance matrices in Figure \ref{fig:dt} are calculated using only the responses of the 17 participants who gave different answers to the most similar ornament and the least similar ornament tasks for each of the ten ornament sets.

From Figure \ref{fig:dt}, one can infer that for the first set, the participants are mostly consistent with each other. It shows most of them think that the Moroccan ornament is most similar to the \textit{cmm (ol)} and least similar to the \textit{cmm (1)}. The answers were least consistent for the seventh set. That is because participants had a hard time choosing between the ornaments of \textit{p3m1} and \textit{p6} groups as the most similar one to the Moroccan ornament.

\begin{figure}[!htb]
	\centering
	
	\includegraphics[width=0.99\linewidth]{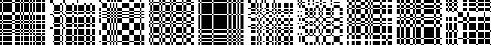}\\
	(a) Questions 1-10\\
	\includegraphics[width=0.99\linewidth]{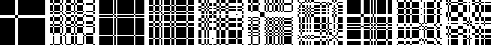}\\
	(b) Questions 11-20\\
	\caption{30x30 distance matrices for twenty task.}\label{fig:dtEach}
\end{figure}

Distance matrices in Figure \ref{fig:dtEach} are calculated using participants' responses for each of the 20 tasks: Identical answers are scored as 0, and different answers are scored as 1. In this case, there was no inconsistency, as each task was evaluated separately; so the answers of all of the 30 participants were included.

Observe that none of the ornaments of $cmm$ group, except for the $cmm$ $(ol)$ was ever selected as the most similar one to the Moroccan ornament. On the contrary, the $cmm$ ornaments were always selected as least similar. In task $3$ (see Fig.~\ref{fig:set1_result}) the participants mostly voted for $p6m$ ornament rather than $cmm$ $(ol)$ ornament as the most similar one to the query. 

\subsection{Second Experiment}

Based on the results of the first experiment we formed a new experiment. Part $a$ and part $c$ (Fig.~\ref{fig:dataset}) from the dataset are used this time. We kept the ornaments that were selected by participants of the first experiment as the most similar ornaments to the Moroccan ornament. Thus, the ornaments of $cmm$ group, other than the $cmm$ $(ol)$ ornament, are discarded since they were selected as least similar to Moroccan ornament. Instead, two new ornaments of $cmm$ group were created, one with no color symmetry, and the other one of $p4g/cmm$ group. As the $cmm$ ornaments in the first experiment contain sparse motifs, where there are many spaces between motifs, for the second survey we decided to create more dense, truly interlocking $cmm$ ornaments. This is done to increase the chances of $cmm$ ornaments to be selected as the most similar one to the Moroccan ornament. We have also modified two ornaments from part $b$ in order to make the symmetries more visible. Thus, the $p6$ $(3)$ and $p6$ $(5)$ ornaments of part $c$ are modified versions of $p6$ $(2)$ and $p6$ $(1)$ ornaments of part $b$. For the first one we made the fact that there is no mirror reflection more visible by swapping colors, and for the second one we made lines bold to stress out six-fold rotations.
\begin{figure}[!htb]
	\centering
	
	\includegraphics[width=0.99\linewidth]{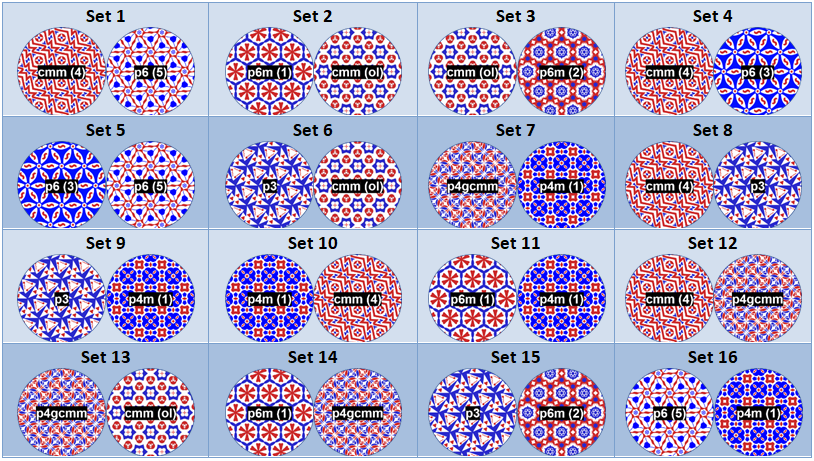}\\
	
	\caption{Ornament sets for the second experiment}\label{fig:set2}
\end{figure}

Overall $16$ ornament sets, each containing two ornaments, were constructed for the second experiment. This time for each set we designed only one task: selecting the ornament which has the same or similar structure to the query ornament. Thus, the number of tasks is also $16$. The query is again the Moroccan ornament in all tasks. The $16$ ornament sets are given in Fig.\ref{fig:set2}.

\begin{figure}[!htb]
	\centering
	
	\includegraphics[width=0.8\linewidth]{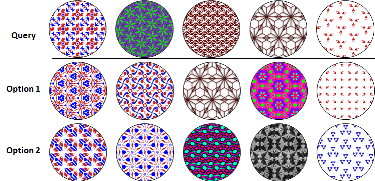}\\
	
	\caption{Five warm up tasks}\label{fig:warmup}
\end{figure}

$20$ participants took part in the second experiment. All of them, except for three, also participated in the first experiment. As before, prior to the beginning of the experiment, we showed some introductory slides, describing the geometric operations and their usage to create an ornament. There were five warm up tasks which are shown in Fig.~\ref{fig:warmup}. One of the options in any warm up task belongs to the same symmetry group as the query ornament. Note that, this is not always true for real experiment tasks, where it might be the case that none of the options belongs to the same symmetry group as the Moroccan ornament. For example, in task $9$ the ornament of $p3$ group and the ornament of $p4m$ group are given as options. Hence, none of these two ornaments match the symmetry group of the query ornament. For the first three tasks in the warm up stage we gave the correct answer only after the participants made their own choice, and they were not allowed to change their answers subsequently.

\begin{figure}[!htb]
	\centering
	\begin{tabular}{cc}
		\includegraphics[width=0.448\linewidth]{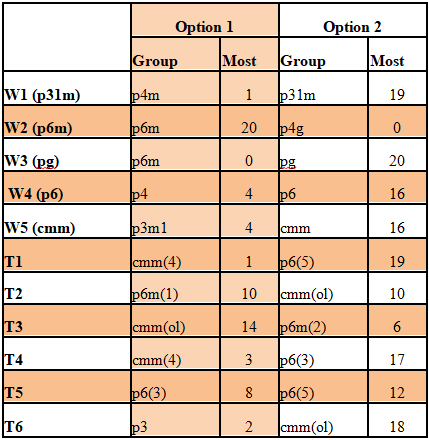}&
		\includegraphics[width=0.45\linewidth]{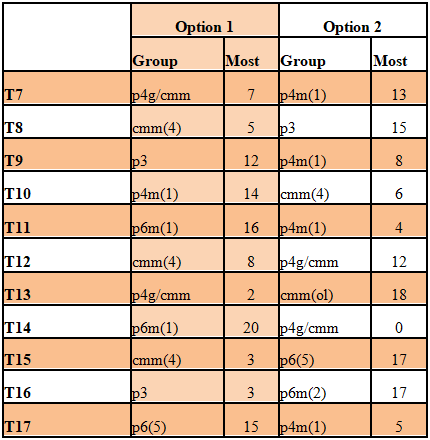}\\
	\end{tabular}
	\caption{Results for the second experiment.}\label{fig:set2_result}
\end{figure}

Fig.~\ref{fig:set2_result} illustrates the results for the second experiment. First five rows of the left table show the answers for the warm up tasks. Observe that the participants mostly selected the correct ornaments in the warm up tasks.

From the experiment results we construct a small sparse matrix, containing the similarities between two ornaments used in the second experiment. We compute it in the following manner. For one particular ornament in a task of the experiment, its similarity to the Moroccan ornament is the ratio of the number of participants that selected this ornament as the most similar one to the overall number of participants ($20$ in our case). If one ornament was used in several tasks then the average of the computed similarity scores is used. The similarity of two optional ornaments in a task is the ratio of the number of participants that selected these ornaments as most similar. For example, the $cmm$ $(ol)$ was used in four tasks. For task $2$ it got $10$ out of $20$. Then for this particular task its similarity to query ornament is $10/20$. But the overall similarity is $(10/20+14/20+18/20+18/20)/4=0.75$. The similarity between $p6m$ $(1)$ and $cmm$ $(ol)$ can be computed from task $2$ as $10/10=1$. If two ornaments were never used in the same task then their similarity is zero. 

\begin{figure}[!htb]
	\centering
	\begin{tabular}{cc}
		\includegraphics[width=0.47\linewidth]{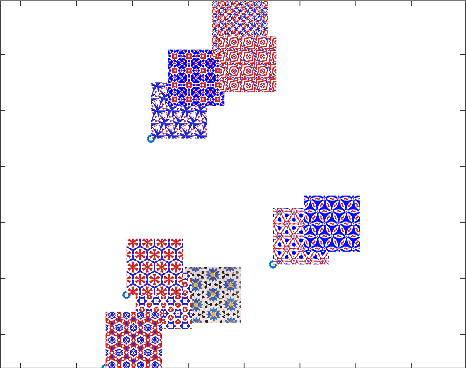}&
		\includegraphics[width=0.45\linewidth]{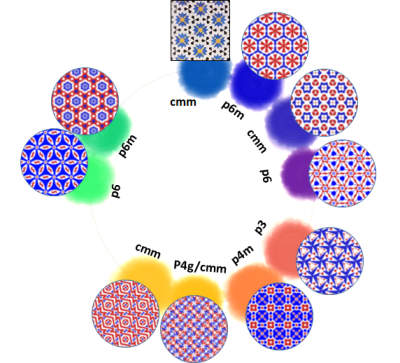}\\
		(a)&(b)\\
	\end{tabular}
	\caption{Visualization of the similarity matrix after reducing it to (a)2D and (b)3D.}\label{fig:clustering}
\end{figure}

In order to visualize the similarities between ornaments we use t-Stochastic Neighbourhood Embedding technique (tSNE) introduced by~\cite{Maaten2008}. tSNE helps to reduce the dimensionality of the similarity matrix, while preserving between object similarities as much as possible. Fig.~\ref{fig:clustering} demonstrates the results after reducing the dimensions of the similarity matrix to two and three. Three dimensional results are mapped to RGB color space, so that the ornaments that are close to each other are represented with the similar or close to similar colors.

Observe from Fig.~\ref{fig:clustering} that, according to the participants, the most similar ornaments to the Moroccan ornament are the the ones with six-fold rotations. Among the ornaments with six-fold rotations, those with mirror reflections are considered to be more close. Also, observe that even though we created $cmm$ $(ol)$ as an exact copy of Moroccan ornament, the participants are mostly indecisive whether the $p6m$ ornament is closer to the query or $cmm$ $(ol)$. While we have fully analysed and proved that the Moroccan ornament belongs to the $cmm$ group, the participants could not see this connection. Moreover, the ornaments of $cmm$ group, except for $cmm$ $(ol)$ are selected as least similar ones.

\section{Conclusion}

Symmetry is a well studied and established topic in theoretical world. However, the classical theory of symmetry groups is incapable of capturing the perception of symmetry. Our work was constructed around a single challenging ornament. First, the Moroccan ornament was analysed within group-theoretical framework. Then two perceptual experiments were conducted where the Moroccan ornament played a leading role. The experiment results showed that the symmetry group of the ornament detected within group-theoretical framework, is selected by participants as the least similar symmetry group. We conclude that what humans see is beyond those discrete and  predefined seventeen symmetry groups. While the definition of symmetry groups lies in small details, humans seem to capture the first striking (which is generally highest) symmetry and then perform comparison between the symmetries of the ornament, without considering the small details that distinguish two symmetry groups.

\bibliographystyle{unsrt}

\begin{thebibliography}{10}
	
	\bibitem{Polya}
	G.~Polya.
	\newblock {\"U}ber die analogie der kristallsymmetrie in der ebene.
	\newblock {\em Zeitschrift f{\"u}r Kristallographie}, 60(1):278--282, 1924.
	
	\bibitem{Niggli24}
	P.~Niggli.
	\newblock Die fl{\"a}chensymmetrien homogener diskontinuen.
	\newblock {\em Zeitschrift f{\"u}r Kristallographie}, 60:283--298, 1924.
	
	\bibitem{Niggli26}
	P.~Niggli.
	\newblock Die regelm{\"a}ssige punktverteilung l{\"a}ngs einer geraden in einer
	ebene.
	\newblock {\em Zeitschrift f{\"u}r Kristallographie}, 63:255--274, 1926.
	
	\bibitem{Adanova2019}
	V.~Adanova and S.~Tari.
	\newblock Analysis of planar ornament patterns via motif asymmetry assumption
	and local connections.
	\newblock {\em Journal of Mathematical Imaging and Vision}, 61(3):269--291,
	2019.
	
	\bibitem{Liu2004_pami}
	Y.~Liu, R.~T. Collins, and Y.~Tsin.
	\newblock A computational model for periodic pattern perception based on frieze
	and wallpaper groups.
	\newblock {\em {IEEE} Transactions on Pattern Analysis and Machine
		Intelligence}, 26(3):354--371, 2004.
	
	\bibitem{Attneave54}
	F.~Attneave.
	\newblock Some informational aspects of visual perception.
	\newblock {\em Psychological Review}, 61(3):183--193, 1954.
	
	\bibitem{Attneave55}
	F.~Attneave.
	\newblock Symmetry, information, and memory for patterns.
	\newblock {\em The {American} Journal of Psychology}, 68(2):209--222, 1955.
	
	\bibitem{washburn1988symmetries}
	D.~K. Washburn and D.~W. Crowe.
	\newblock {\em Symmetries of culture: Theory and practice of plane pattern
		analysis}.
	\newblock University of Washington Press, 1988.
	
	\bibitem{corballis1974perception}
	M.~C. Corballis and C.~E. Roldan.
	\newblock On the perception of symmetrical and repeated patterns.
	\newblock {\em Perception \& Psychophysics}, 16(1):136--142, 1974.
	
	\bibitem{palmer1978orientation}
	S.~E. Palmer and K.~Hemenway.
	\newblock Orientation and symmetry: Effects of multiple, rotational, and near
	symmetries.
	\newblock {\em Journal of Experimental Psychology: Human Perception and
		Performance}, 4(4):691, 1978.
	
	\bibitem{shepard1971mental}
	R.~N. Shepard and J.~Metzler.
	\newblock Mental rotation of three-dimensional objects.
	\newblock {\em Science}, 171(3972):701--703, 1971.
	
	\bibitem{landwehr2011visual}
	K.~Landwehr.
	\newblock Visual discrimination of the 17 plane symmetry groups.
	\newblock {\em Symmetry}, 3(2):207--219, 2011.
	
	\bibitem{Clarke}
	A.~D.~F. Clarke, R.~Green, Patrick, F.~Halley, and M.~J. Chantler.
	\newblock Similar symmetries: The role of wallpaper groups in perceptual
	texture similarity.
	\newblock {\em Symmetry}, 3(2):246--264, June 2011.
	
	\bibitem{kohler2021}
	P.~J. Kohler and A.~D.~F. Clarke.
	\newblock The human visual system preserves the hierarchy of two-dimensional
	pattern regularity.
	\newblock {\em Proceedings of the Royal Society}, 288(1955), 2021.
	
	\bibitem{iornament}
	J.~Richter-Gebert.
	\newblock Science-to-touch.
	\newblock \url{http://www.science-to-touch.com/en/index.html}, 2012.
	
	\bibitem{kendall}
	M.~G. Kendall.
	\newblock A new measure of rank correlation.
	\newblock {\em Biometrika}, 30(1/2):81--93, 1938.
	
	\bibitem{Maaten2008}
	L.~J.~P. van~der Maaten and G.~E. Hinton.
	\newblock Visualizing high-dimensional data using {t-SNE}.
	\newblock {\em Machine Learning Research}, 9:2579--2605, 2008.
	
\end{thebibliography}

\end{document}